\documentclass{article} 
\usepackage{iclr2025_conference, times}

\usepackage{amsmath,amsfonts,bm}

\def\eqref#1{equation~\ref{#1}}

\def\1{\bm{1}}

\DeclareMathAlphabet{\mathsfit}{\encodingdefault}{\sfdefault}{m}{sl}
\SetMathAlphabet{\mathsfit}{bold}{\encodingdefault}{\sfdefault}{bx}{n}

\usepackage{hyperref}
\usepackage{url}

\title{The Philosophical Foundations of Growing AI Like A Child}

\iclrfinalcopy
\author{Dezhi Luo, Yijiang Li, Hokin Deng\\
 University of Michigan, University of California San Diego, Carnegie Mellon University\\ 
\texttt{ihzedoul@umich.edu, yijiangli@ucsd.edu, hokind@andrew.cmu.edu}
}

\begin{document}

\maketitle

\begin{abstract}

Despite excelling in high-level reasoning, current language models lack robustness in real-world scenarios and perform poorly on fundamental problem-solving tasks that are intuitive to humans. This paper argues that both challenges stem from a core discrepancy between human and machine cognitive development. While both systems rely on increasing representational power, the absence of core knowledge—foundational cognitive structures in humans—prevents language models from developing robust, generalizable abilities, where complex skills are grounded in simpler ones within their respective domains. It explores empirical evidence of core knowledge in humans, analyzes why language models fail to acquire it, and argues that this limitation is not an inherent architectural constraint. Finally, it outlines a workable proposal for systematically integrating core knowledge into future multi-modal language models through the large-scale generation of synthetic training data using a cognitive prototyping strategy.

\end{abstract}

\section{Introduction}
Recent advancements in artificial neural networks, particularly transformer-based large language models (LLMs) \citep{brown2020language, touvron2023llama, bai2023qwen, jaech2024openai}, mark perhaps the most impressive progress in artificial intelligence (AI). Never have any AI models performed so well on tasks known to require high-level reasoning abilities, including solving challenging mathematical problems \citep{ahn2024large}, writing code for complex programs that run smoothly \citep{zhuo2024bigcodebench}, drafting travel and business plans \citep{xie2024travelplanner}, and, importantly, doing so entirely through natural language that humans can effortlessly understand. With the emergence of multi-modal language models (MLLMs) \citep{li2023blip2, liu2024visual, team2023gemini}, which align sensory input with language representations, these models show promise for developing intelligence grounded in the physical world and have already demonstrated unprecedented abilities in high-level perception and reasoning \citep{fu2023mme, yang2025magic}. However, despite such astonishing achievements, scientists and users have observed limitations in LLMs that would seem bizarre for humans: their performance on these tasks can drop drastically with slight tweaks to task conditions \citep{yuan2023revisiting,varma2024ravl}.

Said limitations, which concern the robustness of problem-solving, have posed odds over the reliability of deploying LLMs to handle real-life tasks, at least in ideally low-supervised manners \citep{mitchell2023debate}. Interpretations over the underlying causes of such limitations vary, with a prominent account citing the distinction between formal and functional linguistic competences — the ability to produce fluent languages versus the ability to understand and use them in the real world — and arguing that the \textbf{robustness challenge} is due to LLMs possessing the former without the latter \citep{mahowald2024dissociating}. The extreme version of this account would be the one claiming that LLMs lack understanding altogether \citep{deng2025knowing}. In other words, they are “stochastic parrots” that can only solve tasks by abusing spurious correlations in the existing dataset, thereby failing to sensibly answer when the questions are dissimilar enough to what they have been trained on \citep{bender2021dangers}.

At the same time, a surge of attention has been put into benchmarking LLMs: assessing them on large-scale repositories of tasks systematically-developed to target distinct reasoning abilities. A key motivation for such efforts is the notoriously mysterious nature of how LLMs work. It is not apparent, even to engineers who built them, how next-token predictions could enable problem-solving that requires advanced inference. By applying experimental paradigms with controlled designs, researchers are able to differentiate what LLMs can and cannot do under specific task conditions. One particularly staggering kind of result from the benchmarking efforts is the discovery that LLMs could fail miserably at tasks that are easy to humans, despite their high achievements on much harder ones \citep{dentella2024testing, li2025core}. Most interestingly, such a discrepancy could occur between tasks demanding abilities on seemingly the same domain. For example, GPT-4o, a state-of-the-art model open to the public, was shown giving wrong answers to a range of basic mathematical questions, some as simple as 1-digit counting, while crushing advanced problem sets that are challenging even to professional mathematicians \citep{williams2024easy}. 

As counterintuitive as it is, this situation is not unfamiliar to the AI research community. As \cite{karpathy2024tweet} noted, phenomena as such simply reveals that LLMs are not exempt from \textbf{Moravec's Paradox}: tasks that are easy to humans could be extremely difficult to machines and vice versa \citep{moravec1988mind}. Like the robustness challenge discussed above, this generates doubt over whether LLMs truly understand what they are doing and discourages their real-life applications.

This paper does the following:

\begin{enumerate}
    \item First, it suggests that the robustness challenge and the Moravec’s paradox are two sides of the same coin: they both arise from the discrepancy between the cognitive development of humans and machines, which can be summarized as scaling up vs. growing up\footnote[1]{See also \citet{tenenbaum2024scaling}.}.
    \item Second, it argues that the key factor among said discrepancy preventing LLMs to overcome both problems is their absence of what is known as core knowledge. Empirical evidence of humans possessing core knowledge as well as machines lacking thereof is discussed.
    \item Third, it discusses possible interpretations of why core knowledge is not mastered by current LLMs. It is concluded that neither of the likely interpretations designate the current foundational architecture of LLMs to be theoretically incapable of developing core knowledge.
    \item Finally, it outlines a promising pathway to train the next generation LLMs with core knowledge, highlighting the large-scale development of synthetic data using a novel cognitive prototyping strategy.
\end{enumerate}

\section{Scaling Up vs. Growing Up}

While the differences between the computational architectures supporting LLMs and human intelligence are extensively discussed, there has been relatively little attention given to how LLMs and humans differ in their development. In the following section, we discuss how looking at such differences facilitates a unified account of the robustness challenge and the Moravec’s paradox.

A widely held belief in the current AI research community is that changes in LLMs’ reasoning performance can be directly attributed to changes in their scale, as tracked by the number of parameters they have in the neural networks and the size of the dataset they are trained on. This belief, hailed as the \textbf{scaling law}, has largely been supported by empirical results observed throughout LLMs’ progression throughout the past several years, and continues to be the primary strategy adopted by major companies for developing more advanced models \citep{kaplan2020scaling}. In this sense, scaling law has been taken as a theory regarding the nature of LLMs’ cognitive development, that is, increased computational power supporting a domain-general learning mechanism \citep{long2024nativism}. As \citet{sutton2019bitter}, in his seminal "The Bitter Lesson", puts it:

\begin{quotation}
“One thing that should be learned from the bitter lesson is the great power of \textbf{general purpose methods}, of methods that continue to scale with increased computation even as the available computation becomes very great…we should stop trying to find simple ways to think about the contents of minds, such as \textbf{simple ways to think about space, objects, multiple agents, or symmetries}. All these are part of the arbitrary, intrinsically-complex, outside world. They are not what should be built in, as their complexity is endless; instead we should build in only the meta-methods that can find and capture this arbitrary complexity.” (p.2; emphasis added)
\end{quotation}

However, a principal limitation of the scaling law as a theory of cognitive development is that it offers minimal explanatory power: it remains unclear how complex reasoning abilities could simply emerge by increasing a model’s parameters or dataset size. This concern applies not only to hypothetical future capabilities, but also to those already seen in larger models but absent in smaller ones. As \citet{churchland1990neural} emphasized in early connectionist research, epistemological theories must be grounded in empirical data across multiple levels of analysis, not just abstract philosophical reasoning. As such, claiming that further scaling alone will bridge all remaining gaps between LLMs and human cognition without understanding how and why scaling worked in the past amounts to an uninformative a priori assumption with limited scientific merit \citep{marcus2018deep, long2024nativism}. While interest in mechanistic interpretability is growing, the scaling law remains largely unexplained at the algorithmic level \citep{marr1982vision}, raising concerns about its sufficiency as a scientifically-grounded pathway for AI development.

On the other hand, understanding how human cognition develops as we grow up, and eventually reaches the level of adults, where AI also strives to achieve, has been the objective of developmental psychologists over the past several decades \citep{spelke2023precis, tomasello2024agency}. The application of specialized experimental paradigms for probing cognitive abilities have produced a large amount of psychological and behavioral data regarding what children can and cannot do across different age groups, which fertilizes empirically-grounded theories of human cognitive development. 

Among these, many support the idea that human cognition develops along distinct stages marked by the acquisition of previously inaccessible abilities, with more complex abilities grounded on simpler ones \citep{brainerd1978stage, barrouillet2015theories}. Such theories have been referred to as \textbf{stage theories}, which framework is pioneered by the work of Jean Piaget, who defined four distinct stages of development: sensorimotor, preoperational, concrete operational, and formal operational \citep{piaget1952origins}. At each stage, children acquire abilities that are previously unavailable to them, each reshaping their understanding of the world on respective domains. For example, a landmark for the concrete operational stage is that children begin to understand that the quantity of things does not change with how they are arranged \citep{piaget1969child}. Coined as “conservation” by Piaget, this substitutes the naive and rigid understanding of quantity available to the preoperational child, which cannot overcome bias such as the length-equals-number strategy, e.g. thinking that a coin row has more coins than another for being more spreaded out \citep{houde1997numerical}. As the physical prototype for more abstract numerical operations, understanding the law of conservation lays the foundation for learning complex mathematical abilities later in life. Similarly, children’s elimination of egocentrism via the acquisition of basic perspective-taking abilities at a young age paves the way for higher-level theory-of-mind reasoning at an older age, such as understanding others' intentions \citep{piaget1969child}. Looking at a longer time frame, sensorimotor abilities such as object permanence, perceptual constancy, and the sense of continuity found to develop very early in life are indispensable for learning about physics and mechanics in school and beyond \citep{piaget1952origins}.

While many details in Piaget’s theory have been debated following more rigorous empirical investigations, modern literature generally support his idea that children first learn simple abilities to reason about the physical world and develop more complex, abstract abilities on top as they grow up\footnote[2]{Most of the debates have been regarding the specific ages of which children acquire different abilities. The difficulty of reaching consensus across empirical sources have largely led to the abandonment of the “stage” notion even among theorists who support neo-Piagetian approaches (Rochat, 2024). It is, however, important to note that the lack of clearcut age parameters do not go against the overall idea that children transition across a hierarchy of simple to complex abilities throughout development — the transition may just not be in a stage-to-stage fashion where each stage is qualitatively different from another in a definable way.} \citep{barsalou2008grounded, samuelson2000grounding, barsalou2010grounded, pezzulo2013computational}. This claim of a \textbf{grounding} nature of human cognitive development offers critical insights to the limitations in LLMs. 

Foremost, having learned simple abilities prior to complex abilities, humans are not subjected to the Moravec’s Paradox\footnote[3]{Although, strictly speaking, the Moravec’s Paradox is termed so on the basis that human intelligence is held as a standard for comparison.}. Moreover, since there are causal relationship postulated between the primitive, simple abilities and the late-coming complex abilities, the grounding view implies that the absence of simple abilities at an earlier time frame would likely affect the complex abilities later, hence a mechanistic link between the Moravec’s Paradox and the robustness challenge, both of which are observed in LLMs. It therefore may be the case that LLMs fail to generalize across conditions like humans when solving complex reasoning tasks because the skills they employ are not grounded upon simpler abilities on relevant domains. Note that this explanation hinges on the hypothesis that LLMs’ computational architecture shares some aspects with that of humans, at least to the extent that the basic patterns of grounding apply. This may not be the case if human-level intelligence, specifically robust high-level reasoning, is multiple-realizable, which many argued to be likely \citep{bechtel1999multiple, mcgrath2024multiple, li2025core}. Especially, be reminded of scaling law, it is often postulated that LLMs’ cognitive development is based on some kind of domain-general learning mechanism, whereas research into human cognition and its grounding seems to be highly domain-specific. 

Nevertheless, human cognitive development might be more mechanistically similar to that of LLMs than it would appear. It is a long tradition to think that what drives the transitions from simple to complex abilities are highly domain-general. Piaget proposed that cognitive development is fundamentally the project of assimilation and accommodation, which is fitting familiar stimuli to existing cognitive structures and reconstructing them to interpret novel stimuli, respectively. These are essentially data-driven processes that do not select for particular patterns of information. In particular, Piaget suggested that the major transition differentiating the concrete operational stage from the preoperational stage is not the acquisition of specific abilities like conservation and perspective-taking per se, but the ability to perform systematic mental operations over structured mental representations, which in turn support said abilities by enabling functions such as reversibility and simulation \citep{piaget1965number}. In his seminal work "The Language of Thought", \citet{fodor1975language} likewise interpreted that:

\begin{quotation}
“...a reasonable account of the stages of cognitive development could be elaborated by referring to increases in the expressive power of such systems. What I think \textbf{one cannot have, however, is that concept learning provides the mechanisms for the stage-to-stage transitions}. That is, if the child's cognitive development is fundamentally the development of \textbf{increasingly powerful representational/conceptual systems}, then cognitive development cannot be the consequence of concept learning.” (p.89; emphasis added)
\end{quotation}

Fodor notoriously argues for the view that learning is not the matter of acquiring new concepts, but this cannot be taken as to suggest an entirely nativist theory of learning resembling that of Plato \citep{simon1976computer, marcus2018deep}. In concert with Piaget’s theory, what he proposed here is that human cognitive development may be driven by increasingly powerful representational abilities that are capable of supporting better ways of conducting hypothesis-testing, and concept learning is just a consequence of this systematic improvement. Neither Fodor’s nor Piaget’s theory is known to offer an explanation of how this improvement take place, which has largely remained to be an open question, yet it is apparent that it implies a remarkable similarity between the dynamic aspect of human cognitive development and the scaling of LLMs, as increased computational resources essentially support more complex patterns of manipulating representations.

From this perspective, it is therefore reasonable to think about LLMs’ limitations by mapping their differences with human cognition given a possible similarity between the common, foundational mechanism of their development. This is the view that both the robustness challenge and the Moravec’s Paradox are due to a missing link between simple abilities and complex abilities. Further, given that humans supposedly acquire complex abilities not by concept learning but rely on domain-general improvements, it is possible that this missing link is not due to problems with the way of increasing representational powers (i.e. scaling), but the failure to acquire simple abilities in the first place. In the next section, we begin to discuss theories and empirical results supporting this hypothesis, taken from both humans and LLMs.

\section{Core Knowledge}

A large body of works in cognitive science have demonstrated that humans possess a basic understanding of several key domains of the world at a very young age, henceforth \textbf{core knowledge} \citep{spelke2003makes, spelke2007core}. This set of knowledge is generally understood as consisting of simple principles regarding objects, actions, number, space, and social relations, including how they relate to each other. Core knowledge is essentially children’s “developmental start-up software” that enables them to navigate and learn about the rich and dynamic nature of the environment in their early life \citep{lake2017building}. Recent behavioural data have further supported a “child as scientist” proposal, showing that children appear to actively formulate intuitive hypothesis and validate them using these abilities to derive knowledge regarding previously unknown aspects of the world \citep{gopnik1996scientist, schulz2007preschool, gweon2010infants, koksal2018development}. These data, while supporting Fodor’s view that hypothesis-testing plays a foundational role in learning during the early days, highlights the domain-specific nature of the process, during which core knowledge of different aspects of the world is employed by the representational resources of the child. 

It is unclear how core knowledge comes about in the minds of individuals. Neuroimaging studies have suggested that functions within these key domains are likely organized modularly even in a child’s brain, exhibiting specialized neural networks supporting core systems of numerical operations (number), spatial navigation (space), theory-of-mind (social relations), and so on \citep{siegal2002neural, newcombe2004starting, nieder2009representation}. This evidence, paired with behavioural findings showing that these core-level abilities are shared by even the most remote, independently-evolved cultural groups, appears to favor the view that core knowledge is somehow innate. In contrast, others have speculated that it might just be early-developed knowledge instead of being “hardwired” in the brain at birth \citep{carey2011precis}. These discussions constitute a major venue of the modern nativist vs. empiricist debate that we would not go into in detail here. In any case, core knowledge appears so early in childhood that it certainly cannot be the result of language-based learning. In turn, languages are likely a set of complex abilities that is developed upon these simple abilities. This marks a major discrepancy between the growing up of humans and the scaling up of LLMs, which acquire all reasoning abilities through linguistic data. However, it would again be a priori to directly view this as an explanation against LLMs’ possession of simple abilities, which undermines the multiple-realizability of cognition \citep{cao2022multiple}. The rapid development of MLLMs further offers potentials toward this direction by supporting sensory-based learning within language model architectures. 

Whether LLMs could or already possess simple abilities like human core knowledge is fundamentally an empirical question. Benchmarking approaches inspired by experimental paradigms in human research have provided a promising pathway toward answering such questions, in which LLMs are essentially evaluated the same way as human participants in psychology experiments \citep{binz2023using, shiffrin2023probing}. Following such approaches, a large-scale benchmark developed by \citet{li2025core} assessed twelve different cognitive abilities in MLLMs. The abilities tested range from simple, core-level abilities like object permanence and perceptual continuity to complex abilities like tool-using and intentionality understanding, spanning over all core knowledge domains. Tasks in said benchmark consist primarily of image or video-formatted adaptations of classic cognitive tasks used in the developmental psychology literature, such as visual cliff task (spatiality), three-mountain task (visual perspective-taking), and fluid mechanics (mechanical reasoning). Remarkably, they found that MLLMs generally perform much better in complex abilities than simple ones. As an example, state-of-the-art models like GPT-4o, while achieving near-human performance on abilities such as intentionality understanding, a higher-level theory-of-mind ability that typically emerges in humans no earlier than 6–7 years of age, fail catastrophically in simpler abilities like visual perspective-taking, which children are reported to master as early as 36 months of age \citep{moll2011does, linsley20243d, gao2024vision, wang2025vision}. Moreover, further in-depth analyses have shown that model performance on low-level cognitive abilities, such as perspective-taking, does not improve alongside model size \citep{li2025core}. In other words, scaling does not seem to allow MLLMs to become better at such abilities. Said results provide critical empirical support for the hypothesis that scaling up LLMs may not allow them to grasp core knowledge. 

\section{Interpreting Core Knowledge Deficits in LLMs}

Why do large language models (LLMs) fail to acquire core knowledge through scale alone, despite access to vast linguistic datasets and the computational resources to process them? A promising approach to this question lies in systematically comparing foundation models with humans, the only known agents that reliably exhibit core knowledge. By analyzing differences in architectural inductive biases, learning dynamics, and modalities of experience, we can formulate testable hypotheses to explain LLMs’ limitations in this domain. These hypotheses can, in turn, inform the design of future models. If such models demonstrate improved alignment with human-like core knowledge abilities, it would serve as a proof of concept that these limitations are not intrinsic to LLMs, but instead stem from specific design choices or training regimes. Below, we outline three potential interpretations of the reasons underlying core knowledge deficits in MLLMs.

\subsection{Lack of Hardwired Domain-specific Faculties}

To begin with, drawing from a longstanding tradition in nativist epistemology \citep{kant1781critique, locke1824works, chomsky1980rules, cowie2002s}, it has been argued that core knowledge must be grounded in innate, domain-specific cognitive faculties—neural structures that are biologically “hardwired” at birth. Neurobiological evidence indicating the eixstence of specialized cortical regions for processing particular domains such as numerosity, object individuation, and social agency have been cited as supports for this view \citep{spelke2007core}. From this perspective, core knowledge deficits in LLMs may be fundamentally unresolvable, given that their architectures lack such dedicated, domain-specific inductive biases.

However, this argument remains inconclusive at best. First, empirically demonstrating the existence of innate domain-specific mechanisms in humans is notoriously difficult. What appears to be "innate" may instead reflect "early knowledge", namely structured patterns that arise from domain-general learning mechanisms rapidly self-organizing into functionally modular systems upon early environmental exposure, consistent with Piaget’s notion of accommodation \citep{piaget1952origins}. Moreover, even if certain faculties are biologically innate in humans, they may not be computationally necessary for acquiring core knowledge in general. To insist otherwise would be to indicate an exception to the multiple realizability of cognitive functions \citep{bechtel1999multiple, cao2022multiple} and to claim that specific cognitive capacities cannot in principle emerge from domain-general learning—even when scaled up with sufficient data and computation \citep{long2024nativism}.

Another parallel concern along this line has been that pre-linguistic infants appear to lack explicit symbolic systems, and no widely accepted mechanisms explain how domain-specific conceptual representations could form solely through perceptual input \citep{carey2011precis}. At the same time, however, reasoning abilities in LLMs appear to emerge entirely from linguistic generation mechanisms, even for capacities that do not depend on language in humans \citep{han2025learning}. Thus, ruling out the possibility that LLM could have capacity to manipulate and abstract over linguistic symbols that enables the emergence of core knowledge representations through pure language learning would require strong theoretical justification \citep{long2024nativism}.

\subsection{Groundless Learning Trajectory}

Secondly, it could be noted that a central challenge in aligning LLM learning with human cognition lies in the temporal dynamics of data exposure. In human development, cognitive growth unfolds through a structured trajectory: infants begin with limited processing capacities and gradually acquire increasingly abstract representations, scaffolded by embodied interaction and core knowledge primitives \citep{pezzulo2013computational}. This progression allows higher-order reasoning to emerge as a natural extension of early sensorimotor experiences. In contrast, LLMs are exposed to vast amounts of heterogeneous textual data all at once—without any temporal ordering, developmental constraints, or grounded scaffolding. As a result, they process low-level and high-level concepts simultaneously, often without forming the kind of layered, hierarchical conceptual structure that supports flexible and context-sensitive reasoning in humans \citep{mitchell2023debate}. Recognizing this gap has motivated growing interest in curriculum learning, training strategies that introduce concepts in an incremental order from simple to complex, as a way to induce more structured, human-like learning trajectories in artificial systems \citep{liu2025not}.

However, the developmental mismatch in temporal sequencing should not be regarded as an inherent limitation of LLMs. The framework of grounded cognition does not prescribe a fixed developmental trajectory, rather, its central claim is that meaningful connections must be established between perceptual modalities and abstract reasoning. In this light, MLLMs are well-positioned to meet this criterion, by integrating visual, embodied, and linguistic inputs, they can construct grounded symbolic representations that exceed what is achievable through text-only training alone \citep{ma2024babysit, shi2025learning, sheta2025behavioral}. Developmentally structured data, organized to progress from simple to complex, may serve as a valuable complementary strategy in supporting this process. Curriculum learning approaches of this kind have been shown to enhance multimodal learning, especially when combined with large-scale text-based pretraining. This is particularly relevant for training core knowledge representations, which require the alignment of low-level sensorimotor regularities with higher-order symbolic abstraction. Nevertheless, the analogy to human development should not be overstretched, for instance, by assuming that models must be trained exclusively on data comparable to that available to children. LLMs and MLLMs rely on fundamentally different learning faculties, and the goal is not to imitate human developmental biology, but to implement functionally equivalent scaffolding that enables the emergence of structured, generalizable reasoning. From this perspective, the central priority is to foster robust links between multimodal representations and inferential capabilities, not to rigidly follow a human-like learning timeline.

\subsection{Buried Too Deeply}

The third, and perhaps most persuasive, interpretation of core knowledge deficits in MLLMs is that LLMs may, in fact, possess core knowledge but are unable to effectively extract and apply it during reasoning. Since the early days of connectionist research, it has been suggested that artificial neural networks, due to their reliance on distributed representations, inevitably face a challenge: as network size increases through scaling, retrieving knowledge representations for reasoning becomes increasingly difficult and computationally costly \citep{hinton1986learning, clark1992presence}. As Chalmers (1990) remarked:

\begin{quotation}
“Not only is compositional structure encoded implicitly in a pattern of activation, but this implicit structure can be utilized by the familiar connectionist devices of feed-forward/back-propagation in a meaningful way. Such a conclusion is by no means obvious a priori—it might well have turned out that the structure was \textbf{"buried too deeply"} to be directly used, and that all useful processing would have had to proceed first through the step of extraction.” (\cite{chalmers1990syntactic}, p. 60; emphasis added)
\end{quotation}

This concern is particularly salient in the context of core knowledge representations. High-level concepts required for complex tasks—such as recalling a specific historical event or recipe—are often encoded in dense, localized regions of the model’s representational space. Their distinctiveness and clustering make them relatively easy to retrieve. In contrast, low-dimensional concepts like perceptual constancy (i.e., the notion that an object’s identity persists despite changes in sensory input) are diffusely represented across many training examples with similar salience. Because foundational concepts like these emerge from broad contextual regularities rather than being tied to distinct lexical or visual anchors, language models often struggle to isolate and systematically apply them. Although such concepts are implicitly embedded throughout the training distribution, they are rarely reinforced in ways that facilitate robust retrieval or compositional reuse \citep{garrigan2008perceptual, green2024perceptual}. This leads to brittle or shallow reasoning on tasks that require generalization from such principles. As a result, although increased scale may allow core knowledge to be encoded within the distributed weights of the model, this does not ensure that the knowledge is organized in a way that supports reliable inference, as the relevant representations often remain highly dispersed and entangled across the parameter space, hindering the model’s ability to access and deploy them for structured reasoning \citep{shani2023towards}.

However, this interpretation also points to a potential opportunity for building core knowledge in LLMs. If such representations are already implicitly encoded within the model’s parameters, they may not be fundamentally absent but rather latent and insufficiently accessible. The key challenge, then, is not whether core knowledge can be represented in foundation models, but whether it can be activated employed effectively for reasoning. It may be thereby possible to exploit these distributed patterns through targeted training strategies, suggesting that the emergence of core knowledge in LLMs is not inherently unattainable.

\section{Moving Forward: Growing AI Like A Child}

Taken together, these interpretations suggest that while core knowledge may be innate in humans, it may not be necessarily unlearnable in artificial systems \citep{long2024nativism}. Innateness reflects evolved inductive biases, not architectural inevitabilities. Thus, core knowledge might be approximated in LLMs through targeted training that distills structured priors from curated, multi-modal data, making relevant principles salient in their knowledge representations. Rather than mimicking human biology, the goal is to functionally reconstruct core knowledge as a scaffold for structured, commonsense reasoning. For this proposal to be viable, we highlight several key conditions:

\begin{enumerate}
\item \textbf{Sensory Grounding}: models must be trained using multimodal inputs, incorporating both visual and linguistic data to approximate the rich perceptual experience of early human learning.

\item \textbf{Large-scale Exposure}: a sufficiently vast dataset is needed to ensure that core knowledge representations are robust and not lost due to extraction difficulties inherent in distributed neural representations. 

\item \textbf{Confound Minimization}: unlike a child, LLMs are capable of leveraging advanced linguistic processing to access abstract information in their training data, making them prone to shortcut-taking by exploiting spurious correlations \citep{li2025evaluating}. Ensuring that training data is curated to prevent such biases is crucial.

\end{enumerate}

To this end, this paper proposes a practical implementation strategy that satisfies the above criteria: large-scale pretraining on synthetic datasets generated via physics engines, curated through a novel cognitive prototyping strategy. The following section provides a detailed description of the approach and addresses potential concerns.

\subsection{Cognitive Prototype}

To generate training datasets that enhance the saliency of low-level cognitive representations while minimizing environmental confounds, one promising strategy is to directly train MLLMs on machine-adapted cognitive experiments widely used in developmental psychology. The recent translation of these paradigms into machine-readable formats—as exemplified by cognitively inspired benchmarks \citep{binz2023using, li2025core}—provides a robust foundation for constructing such datasets.

Many core cognitive concepts manifest in multiple forms and therefore require diverse experimental paradigms for robust assessment. To systematically structure the training process, we propose a standardized framework called cognitive prototyping—a methodology for curating data centered around individual cognitive abilities. A cognitive prototype includes a precise operationalization of a target concept, based on established cognitive tasks, along with schematic definitions of task-relevant conditions that support controlled variation and generalization. This approach enables large-scale data generation while preserving the conceptual rigor of developmental psychology experiments.

For example, the Three Mountain Task can serve as a prototype for training level-2 visual perspective-taking. In its most classic format, a child observes a three-dimensional model featuring three distinct mountains—one snow-covered, one marked with a red cross, and one topped with a hut—from multiple vantage points. The child is then asked to infer what a doll sees from a different viewpoint by selecting the correct image from a set of photographs \citep{piaget1969child}. This paradigm can be schematized into a range of task-relevant and task-irrelevant variables, such as the spatial positioning and visual appearance of the doll, the observer’s viewpoint offset, the model’s geometry, and the labeling configuration \citep{gao2024vision, li2025core} \footnote[4]{\citet{li2025core} provides additional details on the concept prototyping procedure (see Fig.3). While said discussion focuses on curating evaluation sets, the same abstraction and control framework is equally applicable to constructing training datasets.}. By systematically varying and recombining these parameters, large-scale and diverse training data can be generated while maintaining fidelity to the core conceptual structure of the task. This helps prevent overfitting to superficial cues and encourages models to abstract the underlying principle.

Importantly, cognitive prototypes also support cross-concept generalization within shared environments. For instance, the same Three Mountain setting, when paired with alternative labeling schemes or question types, can be adapted to train level-1 visual perspective-taking or mental rotation \citep{wang2025vision}. This compositional structure allows hierarchically related cognitive abilities to be trained within a unified synthetic environment, enabling models to construct knowledge representations that reflect the interrelatedness of these concepts in the real world \citep{sun2024probing, luo2024vision,  sun2025probing}. By leveraging cognitive prototypes for structured data generation, this strategy offers a scalable and theoretically grounded approach to training MLLMs on core cognitive capacities. It addresses key limitations in current model development by combining large-scale learning with conceptually meaningful scaffolding.

\subsection{Synthetic Data Via Physical Simulations}

Building on the cognitive prototype framework, the next step is to generate large-scale datasets by systematically varying task conditions. Each generated instance represents a distinct combination of task parameters while preserving the underlying cognitive construct specified by the prototype. However, manually labeling such extensive datasets would be infeasible and would severely constrain both scale and diversity \citep{zhang2025can}. To overcome this limitation, we propose leveraging \textbf{synthetic data generation}—a computational strategy for producing richly annotated, scalable, and precisely controlled datasets aligned with cognitive paradigms.

Instead of relying on real-world data, task instances can be rendered using physics engines such as MuJoCo \citep{todorov2012mujoco} or Genesis \citep{zhou2024genesis}, which enable fine-grained manipulation of environmental and agent parameters. These simulation platforms offer the ability to instantiate cognitive experiments with high fidelity and reproducibility while systematically varying key conditions. By embedding task-relevant variables directly into the generation pipeline, such as object positions, agent viewpoints, labeling schemes, and visual occlusions, these engines can produce large volumes of synthetic data that maintain conceptual rigor while avoiding common artifacts and biases found in naturalistic datasets. In combination with cognitive prototypes, this simulation-based approach enables the scalable creation of high-quality training data for MLLMs. It promotes the development of core cognitive abilities by reinforcing structurally meaningful variations, discouraging shortcut learning, and scaffolding representations that mirror grounded knowledge representations in humans.

\subsection{Addressing Concerns: can hollow representations be avoided?}

A salient critique of synthetic training data is that it risks producing "hollow representations": knowledge representation that tailors only to highly rigid scenarios of core cognitive tasks, as opposed to grounded conceptual structures akin to human core knowledge that can be reliably leveraged in high-level reasoning. According to this critique, synthetic training environments, despite their controllability, lack the rich, embodied correspondence between perception and action that grounds human cognition. Consequently, the representations they produce might capture only surface-level statistical features, failing to generalize beyond their designed scope.

However, this concern rests on the assumption that grounding must emerge through human-like experiential data—an assumption we challenge under the “Buried Too Deeply” hypothesis introduced above. Core knowledge is not absent from large-scale pretraining corpora; rather, it is diffusely distributed across the representational manifold, making it difficult for models to access and reuse during reasoning. The limitation, therefore, lies not in the content of representations but in their retrievability and saliency. Knowledge distillation through synthetic data can address this challenge by providing cues for triggering associations between distributed representations through the scaffolding across simple and complex abilities along a given core knowledge domain, increasing their functional accessibility for reasoning. 

This is supported under the standpoint of grounded cognition, robustness arises from grounded representational dependencies—where low-level sensory invariances serve as the substrate for higher-order abstraction. Crucially, this framework does not prescribe a fixed developmental trajectory. Temporal ordering in training—whether mimicking childhood or not—is secondary to associative accessibility: what matters is whether high-level reasoning reliably triggers the relevant low-level conceptual primitives when required. In this light, cognitively structured synthetic data do not aim to replicate human development but to restore associative coupling between representational layers that current MLLMs fail to exploit. Prototype-based synthetic data generation, following a structured hierarchical progression from simple to complex abilities, provides precisely this form of representational scaffolding. By encoding the latent structure of core cognitive concepts within parameterized and compositional synthetic environments, this approach enables the model to align distributed low-level representations with the higher-order inferential routines that depend on them. Such relational strengthening mitigates core knowledge dissociation—the tendency of MLLMs to exhibit symbolic reasoning that is divorced from grounded perception.

That said, our proposal addresses only one viable solution within the current MLLM paradigm. It remains possible, in principle, to construct models that acquire core knowledge through embodied exploration rather than data-driven reconstruction—effectively “raising” a model through interactive experience \citep{al2024project}. Yet such an approach is unlikely to be scalable given the highly distributed, non-interactive nature of present foundation model infrastructure. Under existing architectures, large-scale synthetic data generation remains the most tractable means of making implicit world structure explicit in parameter space, enabling the emergence of robust, grounded representations without requiring human-like developmental trajectories.

\section{Conclusion}

This paper proposes that the robustness challenge and Moravec’s Paradox, two key limitations of current LLMs with significant scientific and practical implications, may be jointly explained by differences in cognitive development between humans and machines. Specifically, these differences are not found within the dynamic process of improving representational power—a mechanism likely shared between scaling up in LLMs and growing up in humans. Instead, they stem from LLMs' absence of core knowledge, a set of foundational cognitive abilities present in humans from early childhood. This core knowledge serves as the basis for gradually acquiring more complex skills over time. Empirical evidence presented in this paper demonstrates that such abilities are indeed missing in current models. Further analysis of the underlying causes of this core knowledge deficit suggests a viable solution: systematically diffusing low-level representations of core cognitive domains during LLM pretraining.

This analysis underscores the need for future research to explore how core knowledge can be effectively incorporated into LLMs. Rather than contradicting the general principle of scaling laws, this perspective challenges the assumption that intelligence can emerge solely from domain-general mechanisms. Just as humans rely on developmental start-up software, LLMs may require structured early training to scaffold their cognitive growth. Encouragingly, given their ability to process vast amounts of high-level information, LLMs may not require innate structures but rather training data that prioritizes salient representations of low-level concepts. Based on said theorization, this paper proposes an engineering solution leveraging large-scale synthetic data generation via simulation engines to systematically generate task scenarios based on developmental psychology paradigms. Importantly, there appears to be no fundamental technical barrier to pretraining core knowledge. The next step is to implement this approach and rigorously evaluate whether it enhances human-like cognitive competence, particularly in real-world robustness. This research agenda can be best summarized as \textit{growing AI like a child—at scale} \footnote[5]{The authors of this article are leading an engineering project on the conviction of this statement, and see http://grow-ai-like-a-child.com/ for details.}.

\bibliography{iclr2025_conference}
\bibliographystyle{iclr2025_conference}

\end{document}